\documentclass{bmvc2k}

\title{Unifying Part Detection and Association for Recurrent Multi-Person Pose Estimation}

\addauthor{Rania Briq}{briq@iai.uni-bonn.de}{1}
\addauthor{Andreas D{\"o}ring}{doering@informatik.uni-bonn.de}{1}
\addauthor{Juergen Gall}{gall@iai.uni-bonn.de}{1}

\addinstitution{
	Computer Vision Group,\\ University of Bonn
}

\runninghead{Briq, D{\"o}ring, Gall}{RMPP}

\def\ie{\emph{i.e}\bmvaOneDot}

\def\etal{\emph{et al}\bmvaOneDot}

\DeclareMathOperator*{\argmax}{\mathrm{argmax}}

\usepackage[]{algorithm2e}
\usepackage{amssymb}
\newcommand{\R}{\mathbb{R}}
\newcommand{\N}{\mathbb{N}}

\SetKwInput{KwInput}{Input}
\SetKwInput{KwOutput}{Output}
\SetKw{kwInit}{Initialize}
\SetKwRepeat{While}{while}

\begin{document}
	
	\maketitle
\begin{abstract}
	We propose a joint model of human joint detection and association for 2D multi-person pose estimation (MPPE). The approach unifies training of joint detection and association without a need for further processing or sophisticated heuristics in order to associate the joints with people individually. The approach consists of two stages, where in the first stage joint detection heatmaps and association features are extracted, and in the second stage, whose input are the extracted features of the first stage, we introduce a recurrent neural network (RNN) which predicts the heatmaps of a single person's joints in each iteration. In addition, the network learns a stopping criterion in order to halt once it has identified all individuals in the image. This approach allowed us to eliminate several heuristic assumptions and parameters needed for association which do not necessarily hold true. Additionally, such an end-to-end approach allows the final objective to be known and directly optimized over during training. We evaluated our model on the challenging MSCOCO dataset and obtained an improvement over the baseline, particularly in challenging scenes with occlusions. 
\end{abstract}

\section{Introduction}
The task of multiperson human pose estimation is defined as the localization of a predefined set of anatomical joints and their association distinctly to individuals. It is an integral task for many applications in computer vision in which humans are active participants. Such scenarios are complicated due to the huge variations in joint articulation, occlusions, close proximity or overlaps of joints due to interactions. Current prevalent approaches are largely based on two paradigms, the first follows a top down design \cite{10.1007/978-3-319-48881-3_44}, in which first a single person is detected and then his/her joints localized. Such approaches are not robust to occlusions or partial visibility of people and inaccurate detections in this stage carry over to the joint localization stage. The second paradigm follows a bottom up approach design in which first a set of joint candidates of all people are collectively identified, and then grouped into poses for each person individually \cite{Cao_2017_CVPR,Pishchulin2016DeepCutJS}. In current state-of-the-art bottom-up approaches, grouping is performed independently from the training, causing the final objective to remain unknown to the training algorithm and therefore optimization is not carried out directly over the objective. Cao \etal \cite{Cao_2017_CVPR} who follow a bottom up approach, train two different branches for the part confidence map and for the association, referred to as Part Affinity Fields, which are used to capture the relationships between the joints and are used as edge weights for optimization in a graph matching problem during inference. Since the formulation results in a k-dimensional matching problem, an NP-hard problem, the problem was solved using greedy relaxation therefore yielding a suboptimal solution. Furthermore, when such heuristics are employed, it remains unclear how the association can be further refined or generalized. Newell \etal \cite{NIPS2017_6822} have devised a loss that tries to guide the network in learning distinct embeddings for each individual. However, these embeddings are not employed for association during training and are only used in a greedy post-processing approach, therefore limiting the network's ability in utilizing its full potential in optimizing over the ultimate objective. In this work, we propose an end-to-end approach, which conflates part detection confidence maps and association features in order to train an additional model responsible for association, making the entire pipeline differentiable. The approach is evaluated on the challenging MSCOCO keypoint dataset ~\cite{lin2014microsoft}. 

\section{Related Work}
Multi-person pose estimation approaches can generally be divided into two categories, namely top-down \cite{Carreira_2016_CVPR, 10.1007/978-3-319-48881-3_44, Chen_2018_CVPR, xiao2018simple, SunXLWang2019, DBLP:conf/iccv/HuangGT17, he2017maskrcnn} and bottom-up methods \cite{Pishchulin2016DeepCutJS, Cao_2017_CVPR, NIPS2017_6822, Iqbal_CVPR2017, DoeringIG18, kocabas18prn, recurrent_staf, DBLP:conf/eccv/PapandreouZCGTM18, fang16rmpe, 45946, DBLP:conf/iccv/HuangGT17}. Methods that follow a top-down design rely on person detectors and estimate the pose for each detected bounding box individually. For that reason the performance of these approaches is tightly coupled with the performance of the underlying person detector. On the other hand, bottom up approaches directly estimate all joint locations in a single run, which have to be assembled into individual poses afterwards.
Several works have proposed to merge the joint estimates into poses by generating a fully-connected graph \cite{Pishchulin2016DeepCutJS, deeper_cut, Iqbal_CVPR2017, insafutdinov17arttrack} based on the joint estimates, and solving a matching problem by utilizing ILP (Integer Linear Programming). A closely related work is \cite{Wang_2018_ECCV} who introduce a minimum weight set packing formulation to solve the association problem of the part detections while applying Nested Benders Decomposition to achieve a more efficient inference time. Other works \cite{Cao_2017_CVPR, NIPS2017_6822, recurrent_staf} predict features such as vector fields \cite{Cao_2017_CVPR, recurrent_staf} which indicate the correspondence between detected joints. This allows reducing the pose assembly into a greedy bipartite graph matching problem as in \cite{Cao_2017_CVPR}.
Approaches in these two paradigms perform the association in a separate stage that does not participate in the training.
In contrast, Kocabas \etal \cite{kocabas18prn} propose to simultaneously predict body joint confidence maps and person bounding boxes. An additional pose residual network (PRN) utilizes this information to directly regress the corresponding joint locations. Carreira \etal \cite{Carreira_2016_CVPR} introduce a corrective iterative feedback method, in which the CNN predicts an additive correction to the current joint estimate rather than directly regressing the keypoint location in the Cartesian representation.  However, part detection based on confidence confidence maps has proven to be more powerful at capturing spatial relationships between the joints \cite{NIPS2017_6822, 10.1007/978-3-319-46484-8_29, Cao_2017_CVPR, Pishchulin2016DeepCutJS, Wang_2018_ECCV} than regression based bottom-up methods.  

Another problem that is related to multi-person pose estimation is the task of instance segmentation.
Romera-Paredes \etal \cite{10.1007/978-3-319-46466-4_19} propose using a recurrent neural network (RNN) for instance segmentation, but without inferring the label of an instance.	
Salvador \etal \cite{salvador2017recurrent} eliminate this shortcoming by introducing an additional branch that predicts the class of each instance.
However, they do not localize the exact joint locations or address the distinct case of people instances which have unique articulation features that can facilitate segmenting instances in complex scenes containing close interactions. 

To the best of our knowledge, there is no bottom-up method that estimates the confidence maps of each person individually.

	\section{Approach}
	\begin{figure*}[t]
		\centering
		\includegraphics[width=10cm]{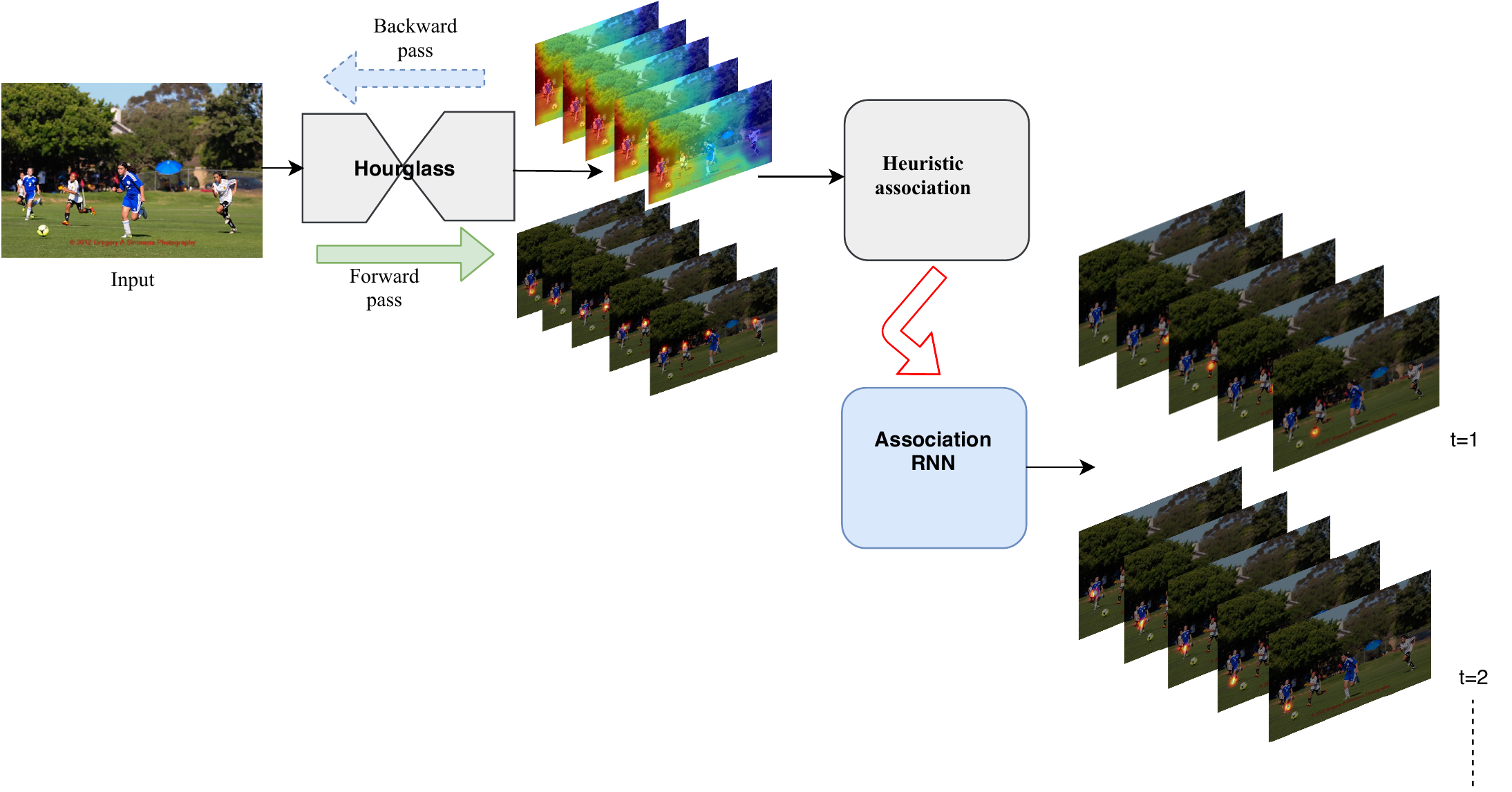}
		\caption{A high level illustration of the approach. The Hourglass network predicts joint heatmaps without differentiating between the different individuals. It follows this stage by a heuristic-based association in which the joints are grouped. The core idea of this work is to replace this association module with an RNN that directly estimates an individual human pose in every iteration.} 
		\label{fig:approach_highlevel}
	\end{figure*} 
	
	\subsection{Associative Embedding Stacked Hourglass}
	The first module in our approach relies on \cite{NIPS2017_6822}. The underlying network in this work is the Stacked Hourglass architecture \cite{10.1007/978-3-319-46484-8_29}, which is a network that stacks together several CNNs that have a symmetric structure where repeated downsampling and upsapmpling are applied in order to capture the full context and refine the predictions. It is designed to capture features at different scales as well in order to predict dense pixel-wise annotations. For pose estimation, its output consists of part detection confidence maps $I\in\R^{J\times m\times m}$, associative embedding maps $E\in\R^{J\times m\times m}$, and the features extracted at the intermediate layers $F\in \R^{f\times m\times m}$, where $J$ is the set of joints, $f$ the number of dense intermediate features and $m$ is the output resolution. Since there are four hourglass stacks, there are four maps for each of these predictions. Each detection heatmap $I_j$ predicts all the joints of type $j$ without distinction as to which person they belong to. In such a design, if there are $k$ people in the image, the heatmap should contain $k$ peaks for every joint $j$. The embedding maps aim at producing similar values for pixels belonging to the same person. The training loss therefore has two terms: the part detection confidence map  $\l_2$ loss and the embedding loss. The embedding loss is devised in such a way that guides the network in producing distinct values for different instances in order to help differentiate between different people in the image. The way the loss is applied does not enforce a specific value, it rather tries to enforce the requirement that the intra-instance distance should be minimized. For further details about how the loss is constructed, we refer to \cite{NIPS2017_6822}.
	
	\subsection{Association Module}
	
	The second module in our approach is responsible for the association and further refinement of the confidence maps, which is intended to eliminate the post processing in \cite{NIPS2017_6822} and make the association as part of the training algorithm. Even though the associative embeddings are good features for separating individuals, the work of \cite{NIPS2017_6822} depends on several post-processing steps, which the RNN eliminates. The operations needed for the grouping are as follows, before finding candidate detections of every joint's confidence map $I_j$, non-maximum suppression is applied. This operation requires specifying a neighborhood size from which the maximum detection is taken, which can erroneously suppress nearby joint detections.
	The candidate detections are thresholded such that only the detections that are above a predefined threshold can be associated with individuals. While iterating over the joint types $j\in J$ in order to decide which person every detection of this type is assigned to, a greedy approach is employed, in which an assignment problem is solved with a cost matrix consisting of the distance between the embeddings corresponding to unassigned detections and the average embedding values of the joints that have been aggregated up until this point. The assignment problem finds a locally optimal choice since it uses only the knowledge about the joints that have been aggregated up until $j$. For associating these joint detections to a person or deciding that a new person is discovered, an embedding threshold value needs to be specified such that only joints within this distance may be grouped together. The authors also favor matching an unassigned joint to a higher confidence detection, \ie if an unassigned joint is close to more than one person by the embedding distance, then it is associated to the person with the highest score. Such a heuristic assumption, while reasonable, does not necessarily always hold true.
	Our proposed approach allows the learning algorithm to optimize over the final grouping without the need for such hand-crafted greedy heuristics with hard assumptions. A high-level overview of the method is described in \ref{fig:approach_highlevel}.
	Invariance to multi-scale transformations remains a challenge for CNNs which are not explicitly designed to be invariant to different scales during training, even when random transformations are applied as part of the data augmentation \cite{NIPS2015_5854}. As such, similarly to \cite{NIPS2017_6822}, we feed forward the image using multiple scales and average the output confidence maps. 
	
\begin{flushright}
	
\end{flushright}	\subsection{RNN as an association module}
	This module contains an RNN network capable of producing a variable output length depending on the input image. In general, RNNs are a helpful model for predicting sequences of variable length with dependency along the sequence. In the case of human pose estimation, the sequence consists of the individuals appearing in the image and the RNN decides where in the image to estimate the next person's pose based on its previous predictions. The RNN should not, for instance, estimate the same person's pose repetitively. Such behaviour is prevented by the memory maintained in the RNN hidden state. In the multiperson pose scenario, the memory implicitly helps in suppressing instances that have been already predicted in previous iterations and allows the network to localize the next person's keypoints in locations that haven't been explored earlier. Similarly to \cite{10.1007/978-3-319-46466-4_19}, the RNN used here is based on Convolutional short-term memory (CONV LSTM)  \cite{NIPS2015_5955}. CONV LSTMs are LSTMs that replace the fully connected layers with convolutional ones in order to attain higher efficiency by better encoding spatial information and by reducing the number of parameters. LSTM contains a memory component that accumulates state information that can be used to optimize over a given objective. 
	
	As input to the RNN, we use the confidence maps, association features referred to as embeddings, and intermediate features which have been extracted from the first module\cite{NIPS2017_6822}. As pointed out earlier, the confidence maps for every joint type are shared for all individuals. In every iteration of the RNN, $J$ confidence maps for a single person are directly inferred.  If $\hat{n}$ people are predicted, then in total $\hat{n}\times J$ confidence maps, denoted by $\hat{H_j}$ are outputted and the RNN is unrolled $\hat{n}$ times before backpropagating the resultant gradient. Each value in the confidence map $\hat{H}$ represents the confidence about the corresponding pixel being the location of joint $j$. The ground truth confidence maps are created with a 2D Gaussian distribution centered at the ground truth location of the joint. Unlike in \cite{NIPS2017_6822}, the confidence maps are created disjointly for each person and are not aggregated into a single confidence map, implying that such a confidence map in the output should contain a single prominent peak corresponding to the joint's location. The number of ground truth confidence maps is therefore equal to $n\times J$, where $n$ is the true number of people in the image. The second value that the RNN predicts is a confidence value $\hat{p}\in[0,1]$ indicating the network's confidence in its current prediction being a new valid pose instance. When this number falls below $0.5$, it indicates that the network believes it has identified all individuals and should halt. The features used to calculate this value are the concatenation of the hidden state matrices from each layer, on top of which a fully connected layer is applied followed by a sigmoid function in order to obtain valid probability values. The ground truth for these values is a vector $p$ that is created as follows:
	\[
	\text{p} =
	\begin{cases}
	\!\begin{aligned}
	& \text{$\underbrace{[1...1}_{n}0]$} \\
	\end{aligned}           & \text{$\hat{n}\leq n$} \\
	\text{$\underbrace{[1...1}_{n}\underbrace{0...0}_{(\hat{n}-n)+1}]$} & \text{otherwise}
	\end{cases}
	\]
	
	the length of the vector $|p|=\max\{\hat{n}, n\}$, where the additional iteration is intended to incorporate knowledge to the network about the stopping criterion. We employ two loss terms on top of the RNN module, the first is the squared $\l_2$ loss between the distinct confidence maps per person and corresponding ground-truth confidence maps.  The second loss, which we refer to as the stopping loss, is calculated using the entropy loss and is responsible for halting the RNN once all the individuals in the image have been identified.   During training, it is possible that the network fails to identify all individuals, \ie $\hat{n}<n$, in which case we let the RNN iterate until $n+1$ but penalize only the first $\hat{n}$ predictions in the confidence map loss. It can be thought of as a way to encourage the network to learn the correct number of people.
	
\begin{figure*}[t]
		\centering
		\includegraphics[width=12cm]{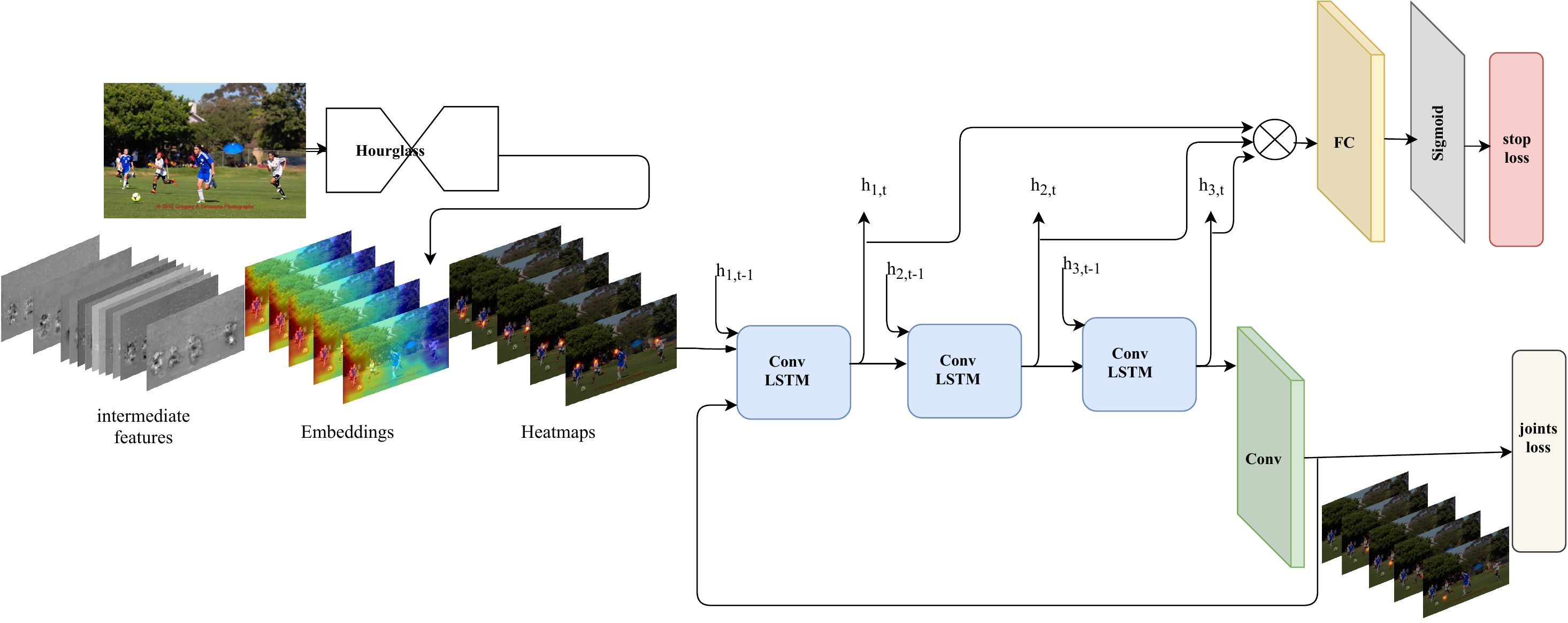}
		\caption{The above diagram presents an overview of our approach. The input to the first ConvLSTM is a concatenation of the current hidden state from the previous iteration and the input features extracted from AE network. The input to the subsequent layers is a concatenation of the output of the previous layer, which has been downsampled by convolution, and the hidden state. The input to the first layer consists of a concatenation of the confidence maps from the hourglass module, the embeddings and the intermediate features, as well as the current RNN prediction in a feedback loop. The features used to calculate the confidence value $p$ are the concatenation of the max pooled hidden state from each layer.
		}
		\label{fig:approach}
	\end{figure*}

	We do not impose any ordering in the output poses and allow the network to reason about which person's joints to estimate next by itself. For computing the confidence map loss, the pose instances naturally need to be associated with the correct individual ground truth confidence map. To that end, we solve an assignment problem using the Hungarian algorithm, in which given the cost matrix $C\in\R^{n\times\hat{n}}$ , the problem requires finding an assignment $S\in\N^{r}$  such that:
	\begin{equation}\label{eq:rnn_loss}
	\sum_{t=1}^{r}\sum_{j=1}^{J}||H_{s_t,j}-\hat{H}_{t,j}||
	\end{equation}
	is minimized, $r=\min\{n, \hat{n}\}$.  Each element in the cost matrix $C$ is given by:
	\begin{equation}\label{eq:rnn_loss}
	C_{kk'}=\sum_{j=1}^{J}||H_{k,j}-\hat{H}_{k,j}||,
	\end{equation}
	which is the sum of distances between each pair of a person's confidence map in the prediction and ground-truth. In summary, the total loss is a sum of the joints loss and stopping loss:
	
	\begin{equation}\label{eq:joint_loss}
	l_{joints} =\sum_{t=1}^{r}\sum_{j=1}^{J}\l_2(H_{s_t,j}, \hat{H}_{t,j})
	\end{equation}
	
	\begin{equation}\label{eq:stop_loss}
	l_{stop}=\sum_{k=1}^{l+1}p_k\log\hat{p_k}+(1-p_k)\log(1-\hat{p_k})
	\end{equation}
	
	\begin{equation}\label{eq:stop_loss}
	L_{total}=l_{joints}+l_{stop}
	\end{equation}
	
	where $l=max\{\hat{n}, n\}$ and both loss terms of the part detections and stop loss are weighted equally. An illustration of the approach is described in figure ~\ref{fig:approach}.

	The value $0.5$ is a reasonable choice for a probability threshold of a positive prediction. To obtain the final joint location $x$, an $\argmax$ operation is applied on each joint's confidence map, \ie $x_j=\argmax{(\hat{H}_J)}$. We specify a threshold value $\tau=0.015$, such that if $\hat{H_j}(x)<\tau$, the joint is discarded. In the multi-scale evaluation, the authors of \cite{NIPS2017_6822} apply single-person refinement for missing joint detection, since we apply the same procedure, to ensure agreement between confidence maps of both outputs, we rather perform $\argmax$ on the multiplication of both confidence maps, \ie $x_j=\argmax{(\hat{H}_J\cdot I_j)}$. 

 Additionally, in order to further refine the  confidence map predictions, we introduce an iterative feedback loop for each person which takes the distinct confidence maps for each person and concatenates them along with the rest of the input features to the RNN in three consecutive iterations. The motivation behind introducing this loop is to encourage the network to produce a single peak in each of its estimation of the disjoint confidence maps. It can therefore be thought of as a means to propagate this constraint in the RNN.
 
	\begin{figure*}[t]
		\centering
		\includegraphics[width=12cm]{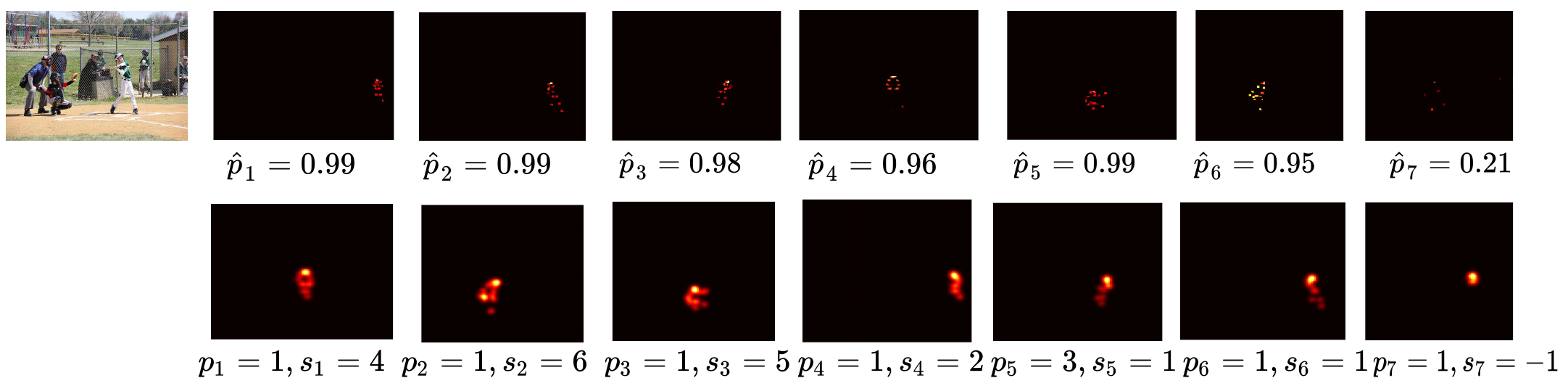}
		\caption{The first raw are the confidence map predictions of the RNN, where each person $i$ is a combination of the $J$ confidence maps predicted at iteration $i$. $s_k=i$ indicates $H_k$ is matched with $\hat{H}_i$. 
			$\hat{p}$ signifies the confidence of the current prediction. The value $p\in\{0,1\}$ is the ground truth confidence values, and $S=\{s\}_{1}^{r}$ is the assignment of the ground truth confidence maps, which appear in some arbitrary order. In this example, the number of people is 8, but the network has managed to identify only six, so no prediction $\hat{H}$ will be assigned to the groundtruth confidence maps $H_7, H_8$ , hence $s_{7,8}=-1$ and they will not participate in the joints loss calculation, but rather only in the stopping loss, where $|p|=9$, and $p_8=1, p_9=0$. }
		\label{fig:assignment}
	\end{figure*}

	\begin{figure*}[t]
		\centering
		\includegraphics[width=14cm]{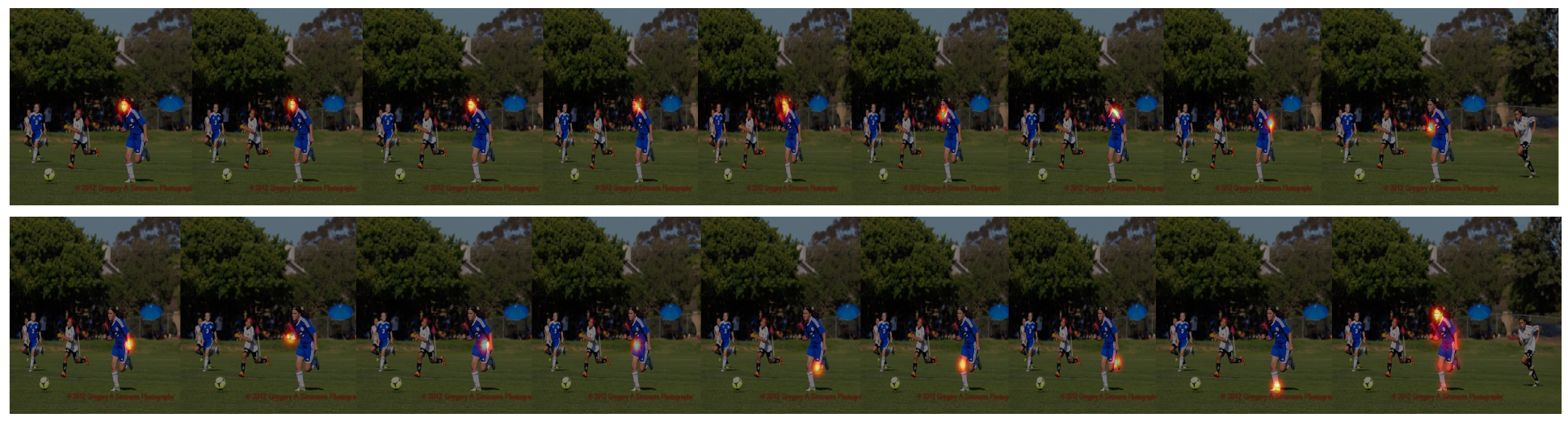}
		\caption{In this plot, we show the output of the RNN association module for a single person, which corresponds to one iteration of the RNN. The last image on the bottom right corner is the sum of all confidence maps and is added for illustration only. As can be observed, every confidence map contains a joint prediction of a single person.}
		\label{fig:disjoint_heatmap}
	\end{figure*} 
	
	\section{Implementation details}
	
\begin{figure*}[t]
		\centering
		\includegraphics[width=8cm]{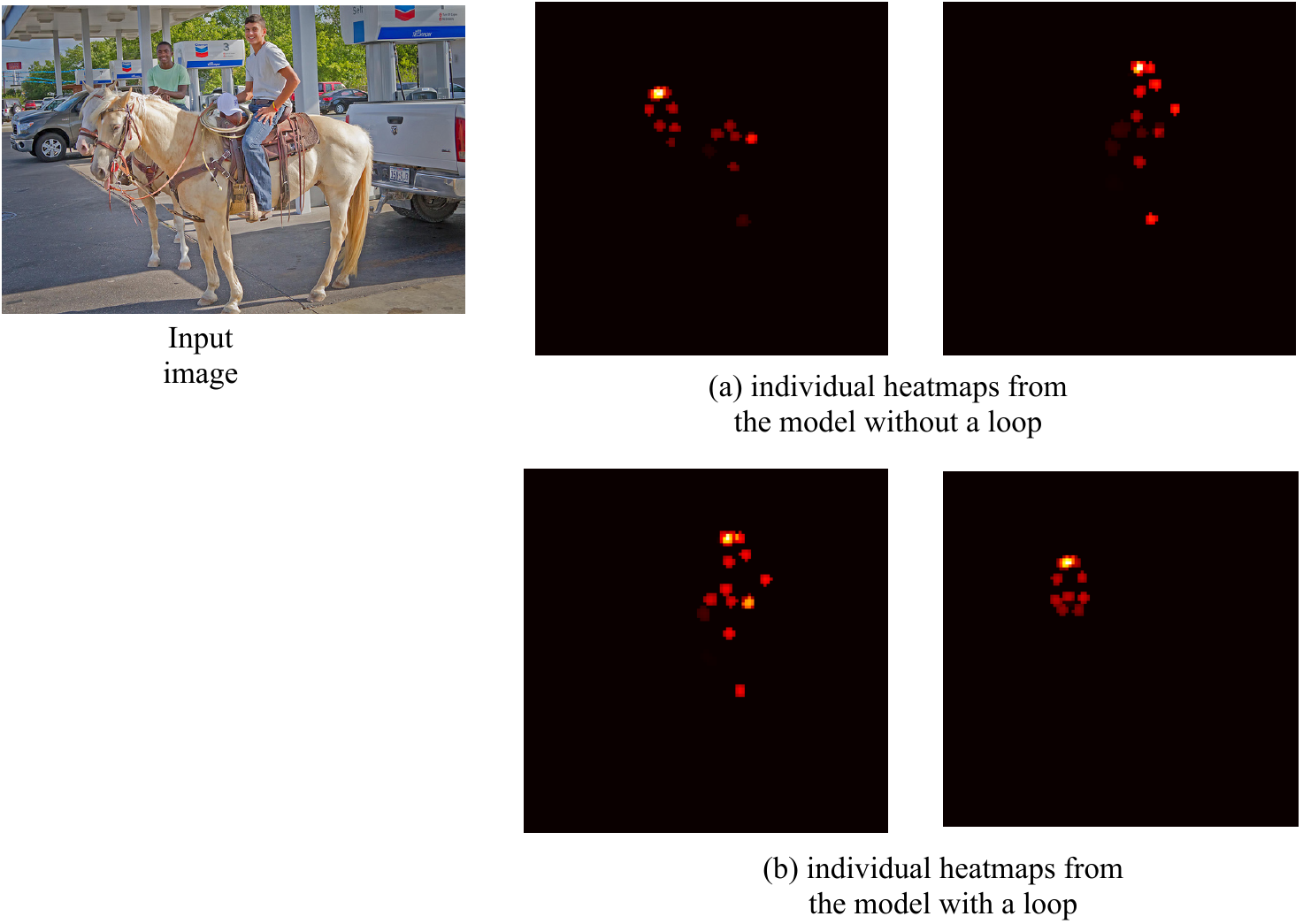}
		\caption{The difference between the result of a model trained without an iterative loop in the first row, and one trained with a loop in the second row. As expected, the the iterative loop helped in suppressing wrong peaks in its confidence map. This can be observed in top left photo, in which the person riding in the back is not distinctly detected from the second person in the front.} 
		\label{fig:iter_illustration}
	\end{figure*}

	Since we rely on the features of the approach in \cite{NIPS2017_6822}, we build our approach on top of their publicly available library. The input to this first module is an image resized to a $512\times512$ resolution. The output of this first network is a set of maps $M \in \R^{4\times64\times 128\times 128}$, where 4 corresponds to the number of Hourglass stacks, and 64 are the channels for the confidence maps whose number is $J=17$, embeddings with the same size as the confidence maps, and intermediate features of size $f=32$. The RNN model used in this work is based on ConvLSTM as presented in ~\cite{NIPS2015_5955}. The kernel size of the convolution in the ConvLSTM is set to 3 and the number of hidden layers in the ConvLSTM is 3. In the first iteration, the confidence maps to the feedback loop are initialized with a uniform distribution $[0,0.1]$. During inference, the network is insensitive to the initialization so we simply initialize them with zero. For examining the influence of this loop, we draw a comparison with a model that has been trained without the loop and observe that indeed the network learns to suppress redundant peaks in the output. The difference is illustrated visually using the confidence maps in figure \ref{fig:iter_illustration}.  
	The network had been trained for five epochs with batch size $b=1$. Due to RNNs huge memory usage, we could not afford to increase it. We used the Adam optimizer, with an initial learning rate of $1e^{-4}$, dropped to $1e^{-5}$ at epoch 4. Constrained by memory limitations, during training, the maximum number of iterations allowed per image was 6, i.e., the network will not be able to iterate for more than 6 iterations, which corresponds to 6 poses, during training. However, since random cropping is part of the data augmentation employed, there is a probability that persons that were missed in the current epoch will be captured in later epochs. We use the  confidence maps of the Hourglass networks from all four stacks and embeddings of the last module only. With respect to the embeddings, one operation to preserve their uniqueness is to concatenate them. Due to memory limitations, we took the embeddings of the last hourglass stack. During inference, memory is not a limitation and the number of maximum iterations allowed is much higher, since no memory is needed for the backpropagation, which constitutes the bottleneck when training an RNN. 
	The code will be made publicly available upon acceptance.

\section{Results and Analysis}
\begin{table*}[t]
	
	\begin{tabular}[h]{lcccc}
		\hline
		Method              & AP &$ AP^{50}$  &$ AP^{M}$ &  $ AP^{L}$ \\
		\hline
		&  & Bottom-Up \\
		\hline
		OpenPose\cite{Cao_2017_CVPR} & 61.8 & 84.9 & 57.1 & 68.2  \\
		PersonLab\cite{DBLP:conf/eccv/PapandreouZCGTM18} & 68.7 & 89.0 & 64.1 & 75.5 \\
		MPNet\cite{kocabas18prn} & 69.6 & 86.3 & 65.0 & 76.3 \\
		
		& &  &\\
		AE singlescale\cite{NIPS2017_6822}  & 56.6 & 81.8 & 49.8 & 67.0  \\
		Ours singlescale  & 57.7 & 81.3 & 48.6 & 71.6\\
		AE multiscale \cite{NIPS2017_6822}  & 65.5 & 86.8 & 60.6 & 72.6  \\
		Ours multiscale & 66.3 & 84.2 & 59.2 & 77.1 \\
		\hline
		& & Top-Down \\
		\hline
		Mask-RCNN \cite{he2017maskrcnn} &  63.1 & 87.3 & 57.8 & 71.4 \\
		G-RMI \cite{45946} & 64.9 &  85.5 & 62.3& 70.0 \\
		CPN \cite{Chen_2018_CVPR} & 72.1 & 91.4 & 68.7 & 77.2 \\
		RMPE \cite{fang16rmpe} &  72.3 & 89.2 & 68.0 & 78.6 \\
		CFN \cite{DBLP:conf/iccv/HuangGT17} & 72.6& 86.1 & 78.3& 64.1\\
		MSRA \cite{xiao2018simple} & 73.7 & 91.9 & 70.3 & 80.0 \\
		HRNet \cite{SunXLWang2019} & 75.5 & 92.5 & 71.9 & 81.5 \\
		\hline
		
	\end{tabular}
	\caption{Results on MS Coco validation set.} \label{tab:val_evaluation}
\end{table*}

In the results in table~\ref{tab:val_evaluation}, we compare with the results of \cite{NIPS2017_6822}, the baseline on which we rely for the feature extraction. We were able to replace their association heuristic with a trained model and further improve the results. 

\begin{figure*}[t]
	\centering
	\setlength\tabcolsep{1pt}
	\scalebox{0.3}{
		\begin{tabular}{llllll}
			\multicolumn{1}{c}{\includegraphics[height=6cm, width=6cm]{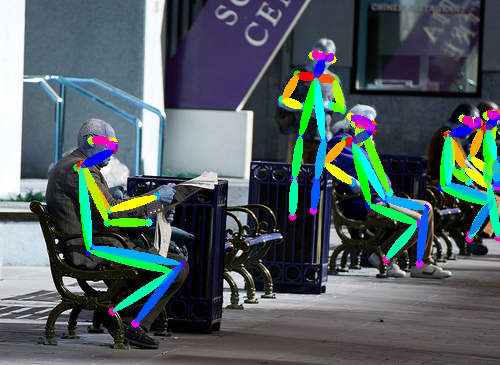}}&
			\multicolumn{1}{c}{\includegraphics[height=6cm, width=6cm ]{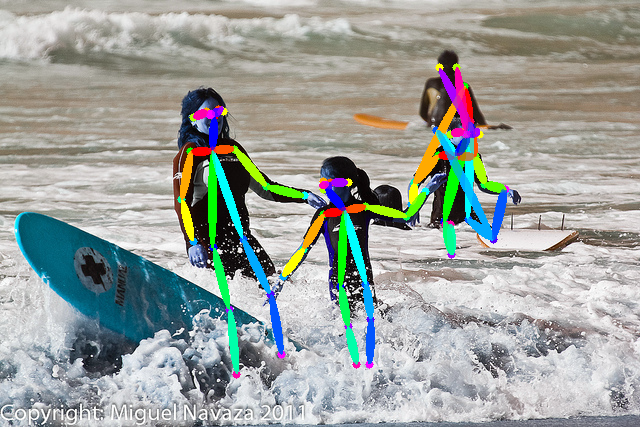}}&  
	 
			\multicolumn{1}{c}{\includegraphics[height=6cm, width=6cm]{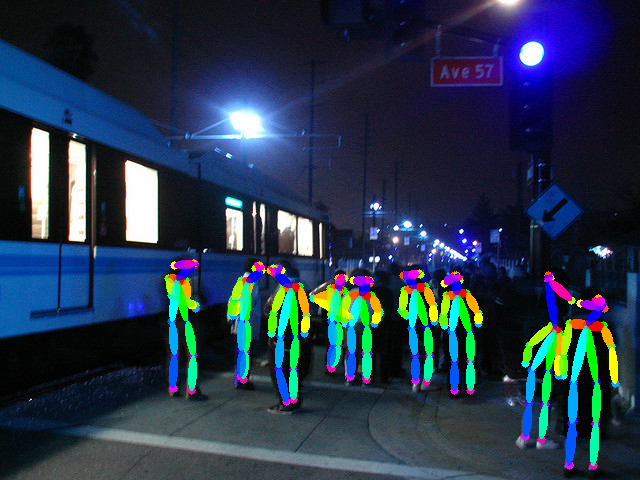}}&\\  
			\multicolumn{1}{c}{\includegraphics[height=6cm, width=6cm]{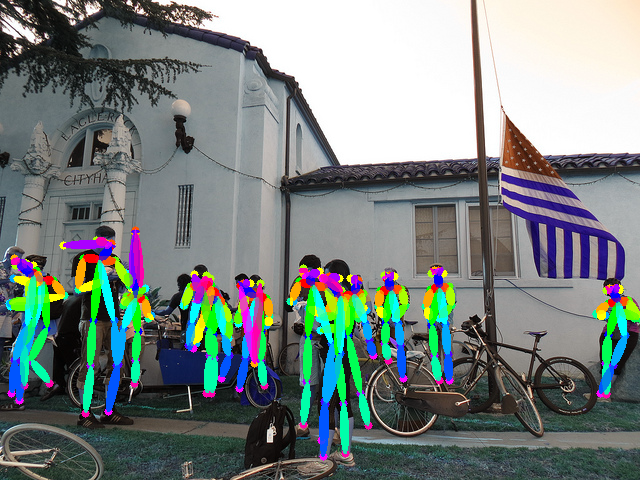}}&  
			\multicolumn{1}{c}{\includegraphics[height=6cm, width=6cm]{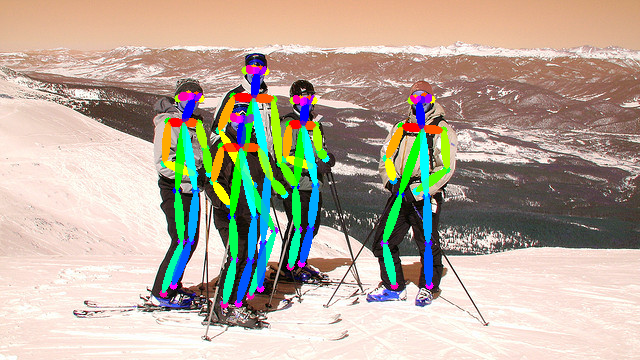}}&

			\multicolumn{1}{c}{\includegraphics[height=6cm, width=6cm]{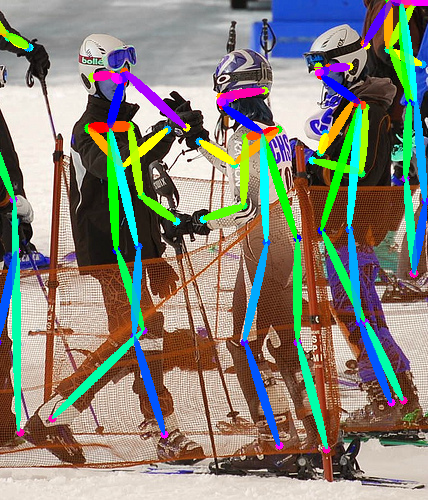}}&

		\end{tabular}}
		\caption{Qualitative results from MSCOCO validation dataset. Failure cases include occluded joints.}
		\label{fig:qualitivate_res}
	\end{figure*}
	In \ref{fig:qualitivate_res} we visualize poses for several images in challenging scenarios. This includes unusual poses, crowded scenes, close interactions and overlaps. The RNN learned to separate individuals well in general. Failure cases include occluded joints, which require separate handling. 
	
\subsection{Comparison with the baseline}
In addition to achieving higher precision compared to the baseline, in figure \ref{fig:qualitivate_res}, we visually demonstrate the advantage of our learning-based association approach compared to the heuristic-based Associative Embedding (AE). We observe that in more challenging scenarios, the RNN performs better at localizing joints and associating them. Such scenarios include overlapping people or occluded joints, in which the RNN is more capable of reasoning about the occluded joints' locations. Additional challenging scenarios include crowded images with a cluttered background. False positives such as statues, human-shaped objects  or people appearing in photos remain a challenge, but in two examples in figure \ref{fig:qualitivate_res}, we observe that the RNN suppressed false positives in which AE could not, making the RNN more robust to such outliers.

In determining the number of people present in the image, AE relies on the distance between the current pose's average embedding and the new joint's embedding value in order to decide whether this new joint should belong to a current pose or a new pose. Oftentimes AE tends to overestimate the number of people in the image resulting in redundant poses. In contrast, the RNN's stopping criterion yields a better estimate of the number of people.

\begin{figure*}[t]
	\centering
	\setlength\tabcolsep{6pt}
	\scalebox{0.4}{
		\begin{tabular}{llllll}
			\multicolumn{1}{c}{\includegraphics[height=7cm, width=7cm ]{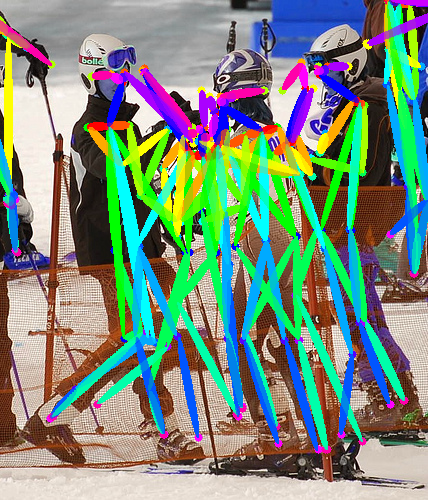}}&
			\multicolumn{1}{c}{\includegraphics[height=7cm, width=7cm]{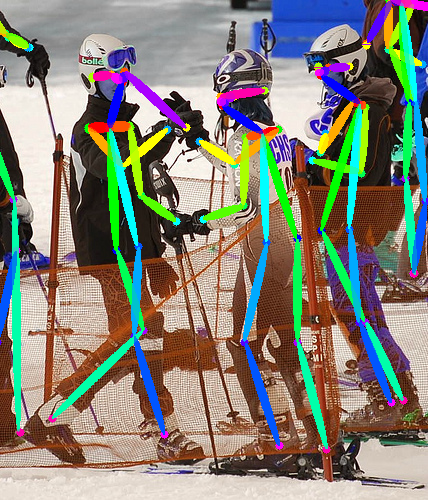}}&
			
			\multicolumn{1}{c}{\includegraphics[height=7cm, width=7cm]{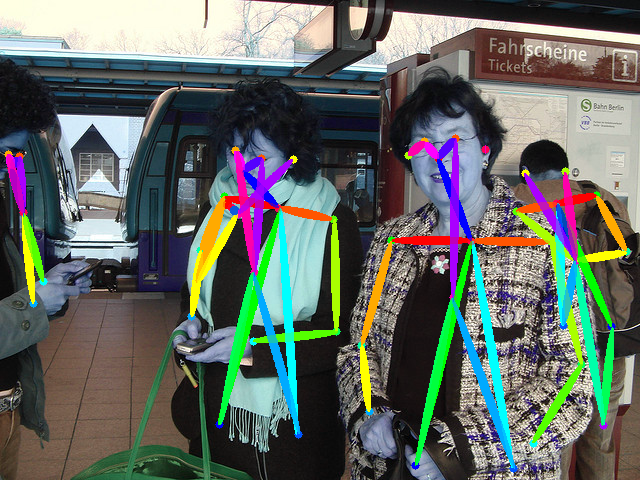}}&  
			\multicolumn{1}{c}{\includegraphics[height=7cm, width=7cm]{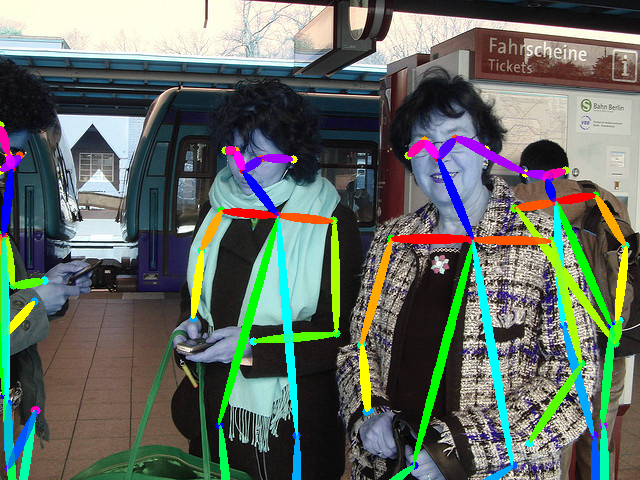}}&\\
			
	\multicolumn{1}{c}{\includegraphics[height=7cm, width=7cm ]{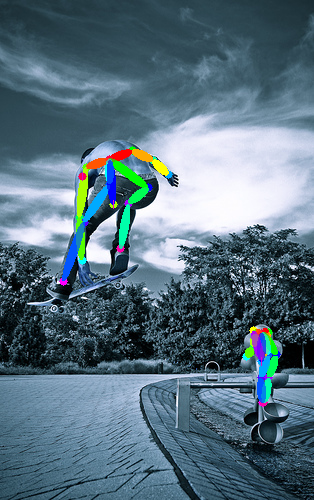}}&
			\multicolumn{1}{c}{\includegraphics[height=7cm, width=7cm ]{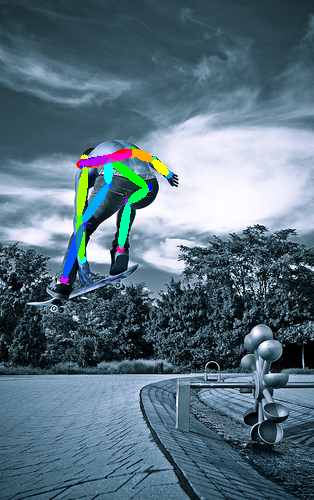}}&
			\multicolumn{1}{c}{\includegraphics[height=7cm, width=7cm]{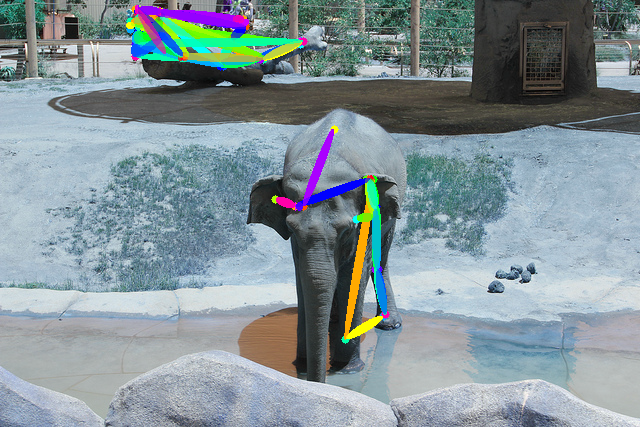}}& 
			\multicolumn{1}{c}{\includegraphics[height=7cm, width=7cm]{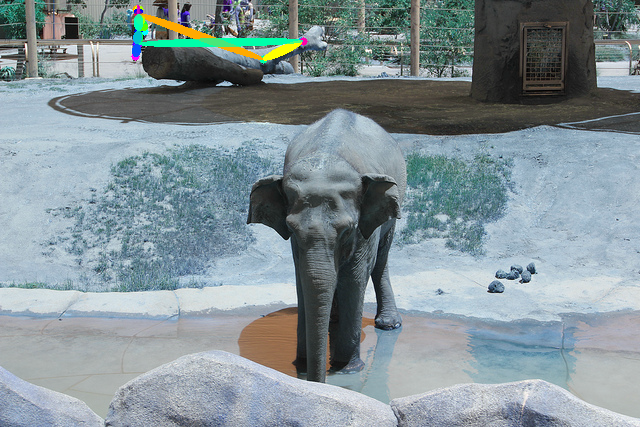}}&\\ 
			\\
			
			\multicolumn{1}{c}{AE}&  \multicolumn{1}{c}{Ours}& 
			\multicolumn{1}{c}{AE}&
			\multicolumn{1}{c}{Ours}& 
			
		\end{tabular}}
		\caption{Qualitative comparison between Associative Embedding and our RNN-based approach.}
		\label{fig:qualitivate_res}
	\end{figure*}
	\clearpage
	\vspace{-4mm}
	
	\subsection{Ablation studies}
	We experimented with different number of loops in the feedback loop and found that the number three produces the best results, as the network started overfitting with more loops. A comparison in the performance can be found in \ref{tab:loop}.
	\begin{table}[t]
		\centering
		\begin{tabular}{lccc|c}
			\hline
			iterations number	& 0 & 3 & 4 \\
			\hline
			AP & 51.5 & 57.7 & 53.3\\
			\hline
		\end{tabular}
		\caption{Examining the influence of the number of iterations in the feedback loop. 0 means without the feedback loop.} \label{tab:loop}
	\end{table}
	
	It is noteworthy that while the part detections heatmaps and associative embedding features are fed as part of the input to the RNN, we found that they are not crucial for the learning algorithm and not having them causes the accuracy to drop by $1\%$ 
	\section{Conclusion}
	Our goal was to make association of human poses part of the learning procedure. The joint model that we proposed yielded results that outperformed the baseline. This proves the importance of making the final objective known and optimized over during training.

	\bibliography{egbib}

\begin{thebibliography}{26}
\providecommand{\natexlab}[1]{#1}
\providecommand{\url}[1]{\texttt{#1}}
\expandafter\ifx\csname urlstyle\endcsname\relax
  \providecommand{\doi}[1]{doi: #1}\else
  \providecommand{\doi}{doi: \begingroup \urlstyle{rm}\Url}\fi

\bibitem[dee(2016)]{deeper_cut}
\emph{DeeperCut: A Deeper, Stronger, and Faster Multi-Person Pose Estimation
  Model}, 2016.

\bibitem[Cao et~al.(2017)Cao, Simon, Wei, and Sheikh]{Cao_2017_CVPR}
Zhe Cao, Tomas Simon, Shih-En Wei, and Yaser Sheikh.
\newblock Realtime multi-person 2d pose estimation using part affinity fields.
\newblock In \emph{The IEEE Conference on Computer Vision and Pattern
  Recognition (CVPR)}, July 2017.

\bibitem[Carreira et~al.(2016)Carreira, Agrawal, Fragkiadaki, and
  Malik]{Carreira_2016_CVPR}
Joao Carreira, Pulkit Agrawal, Katerina Fragkiadaki, and Jitendra Malik.
\newblock Human pose estimation with iterative error feedback.
\newblock In \emph{The IEEE Conference on Computer Vision and Pattern
  Recognition (CVPR)}, June 2016.

\bibitem[Chen et~al.()Chen, Wang, Peng, Zhang, Yu, and Sun]{Chen_2018_CVPR}
Yilun Chen, Zhicheng Wang, Yuxiang Peng, Zhiqiang Zhang, Gang Yu, and Jian Sun.
\newblock Cascaded pyramid network for multi-person pose estimation.
\newblock In \emph{CVPR}.

\bibitem[Doering et~al.(2018)Doering, Iqbal, and Gall]{DoeringIG18}
Andreas Doering, Umar Iqbal, and J{\"{u}}rgen Gall.
\newblock Jointflow: Temporal flow fields for multi person pose estimation.
\newblock In \emph{BMVC}, 2018.

\bibitem[Haoshu~Fang and Lu(2017)]{fang16rmpe}
Shuqin~Xie Haoshu~Fang and Cewu Lu.
\newblock {RMPE}: Regional multi-person pose estimation.
\newblock \emph{ICCV}, 2017.

\bibitem[He et~al.(2017)He, Gkioxari, Doll\'{a}r, and Girshick]{he2017maskrcnn}
Kaiming He, Georgia Gkioxari, Piotr Doll\'{a}r, and Ross Girshick.
\newblock {Mask R-CNN}.
\newblock In \emph{{ICCV}}, 2017.

\bibitem[Huang et~al.(2017)Huang, Gong, and Tao]{DBLP:conf/iccv/HuangGT17}
Shaoli Huang, Mingming Gong, and Dacheng Tao.
\newblock A coarse-fine network for keypoint localization.
\newblock In \emph{ICCV}, 2017.

\bibitem[Insafutdinov et~al.(2017)Insafutdinov, Andriluka, Pishchulin, Tang,
  Levinkov, Andres, and Schiele]{insafutdinov17arttrack}
Eldar Insafutdinov, Mykhaylo Andriluka, Leonid Pishchulin, Siyu Tang, Evgeny
  Levinkov, Bjoern Andres, and Bernt Schiele.
\newblock {ArtTrack: Articulated Multi-person Tracking in the Wild}.
\newblock In \emph{CVPR}, 2017.

\bibitem[Iqbal and Gall(2016)]{10.1007/978-3-319-48881-3_44}
Umar Iqbal and Juergen Gall.
\newblock Multi-person pose estimation with local joint-to-person associations.
\newblock In Gang Hua and Herv{\'e} J{\'e}gou, editors, \emph{Computer Vision
  -- ECCV 2016 Workshops}, pages 627--642, Cham, 2016. Springer International
  Publishing.
\newblock ISBN 978-3-319-48881-3.

\bibitem[Iqbal et~al.(2017)Iqbal, Milan, and Gall]{Iqbal_CVPR2017}
Umar Iqbal, Anton Milan, and Juergen Gall.
\newblock Posetrack: Joint multi-person pose estimation and tracking.
\newblock In \emph{IEEE Conference on Computer Vision and Pattern Recognition
  (CVPR)}, 2017.

\bibitem[Jaderberg et~al.(2015)Jaderberg, Simonyan, Zisserman, and
  kavukcuoglu]{NIPS2015_5854}
Max Jaderberg, Karen Simonyan, Andrew Zisserman, and koray kavukcuoglu.
\newblock Spatial transformer networks.
\newblock In C.~Cortes, N.~D. Lawrence, D.~D. Lee, M.~Sugiyama, and R.~Garnett,
  editors, \emph{Advances in Neural Information Processing Systems 28}, pages
  2017--2025. Curran Associates, Inc., 2015.
\newblock URL
  \url{http://papers.nips.cc/paper/5854-spatial-transformer-networks.pdf}.

\bibitem[Ke~Sun and Wang(2019)]{SunXLWang2019}
Dong~Liu Ke~Sun, Bin~Xiao and Jingdong Wang.
\newblock Deep high-resolution representation learning for human pose
  estimation.
\newblock In \emph{arXiv-preprint}, 2019.

\bibitem[Kocabas et~al.(2018)Kocabas, Karagoz, and Akbas]{kocabas18prn}
Muhammed Kocabas, Salih Karagoz, and Emre Akbas.
\newblock Multi{P}ose{N}et: Fast multi-person pose estimation using pose
  residual network.
\newblock In \emph{ECCV}, 2018.

\bibitem[Lin et~al.(2014)Lin, Maire, Belongie, Hays, Perona, Ramanan,
  Doll{\'a}r, and Zitnick]{lin2014microsoft}
Tsung-Yi Lin, Michael Maire, Serge Belongie, James Hays, Pietro Perona, Deva
  Ramanan, Piotr Doll{\'a}r, and C~Lawrence Zitnick.
\newblock Microsoft coco: Common objects in context.
\newblock In \emph{European Conference on Computer Vision}, pages 740--755.
  Springer, 2014.

\bibitem[Newell et~al.(2016)Newell, Yang, and
  Deng]{10.1007/978-3-319-46484-8_29}
Alejandro Newell, Kaiyu Yang, and Jia Deng.
\newblock Stacked hourglass networks for human pose estimation.
\newblock In Bastian Leibe, Jiri Matas, Nicu Sebe, and Max Welling, editors,
  \emph{Computer Vision -- ECCV 2016}, pages 483--499. Springer International
  Publishing, 2016.
\newblock ISBN 978-3-319-46484-8.

\bibitem[Newell et~al.(2017)Newell, Huang, and Deng]{NIPS2017_6822}
Alejandro Newell, Zhiao Huang, and Jia Deng.
\newblock Associative embedding: End-to-end learning for joint detection and
  grouping.
\newblock In I.~Guyon, U.~V. Luxburg, S.~Bengio, H.~Wallach, R.~Fergus,
  S.~Vishwanathan, and R.~Garnett, editors, \emph{Advances in Neural
  Information Processing Systems 30}, pages 2277--2287. Curran Associates,
  Inc., 2017.
\newblock URL
  \url{http://papers.nips.cc/paper/6822-associative-embedding-end-to-end-learning-for-joint-detection-and-grouping.pdf}.

\bibitem[Papandreou et~al.(2017)Papandreou, Zhu, Kanazawa, Toshev, Tompson,
  Bregler, and Murphy]{45946}
George Papandreou, Tyler Zhu, Nori Kanazawa, Alexander Toshev, Jonathan
  Tompson, Chris Bregler, and Kevin Murphy.
\newblock Towards accurate multi-person pose estimation in the wild.
\newblock 2017.

\bibitem[Papandreou et~al.(2018)Papandreou, Zhu, Chen, Gidaris, Tompson, and
  Murphy]{DBLP:conf/eccv/PapandreouZCGTM18}
George Papandreou, Tyler Zhu, Liang{-}Chieh Chen, Spyros Gidaris, Jonathan
  Tompson, and Kevin Murphy.
\newblock Personlab: Person pose estimation and instance segmentation with a
  bottom-up, part-based, geometric embedding model.
\newblock In \emph{ECCV}, 2018.

\bibitem[Pishchulin et~al.(2016)Pishchulin, Insafutdinov, Tang, Andres,
  Andriluka, Gehler, and Schiele]{Pishchulin2016DeepCutJS}
Leonid Pishchulin, Eldar Insafutdinov, Siyu Tang, Bjoern Andres, Mykhaylo
  Andriluka, Peter~V. Gehler, and Bernt Schiele.
\newblock Deepcut: Joint subset partition and labeling for multi person pose
  estimation.
\newblock \emph{2016 IEEE Conference on Computer Vision and Pattern Recognition
  (CVPR)}, pages 4929--4937, 2016.

\bibitem[Raaj et~al.(2018)Raaj, Idrees, Hidalgo, and Sheikh]{recurrent_staf}
Yaadhav Raaj, Haroon Idrees, Gines Hidalgo, and Yaser Sheikh.
\newblock Efficient online multi-person 2d pose tracking with recurrent
  spatio-temporal affinity fields.
\newblock \emph{arXiv-preprint}, 2018.

\bibitem[Romera-Paredes and Torr(2016)]{10.1007/978-3-319-46466-4_19}
Bernardino Romera-Paredes and Philip Hilaire~Sean Torr.
\newblock Recurrent instance segmentation.
\newblock In Bastian Leibe, Jiri Matas, Nicu Sebe, and Max Welling, editors,
  \emph{Computer Vision -- ECCV 2016}, pages 312--329, Cham, 2016. Springer
  International Publishing.
\newblock ISBN 978-3-319-46466-4.

\bibitem[Salvador et~al.(2017)Salvador, Bellver, Campos, Baradad, Marques,
  Torres, and Giro-i Nieto]{salvador2017recurrent}
Amaia Salvador, Miriam Bellver, Victor Campos, Manel Baradad, Ferran Marques,
  Jordi Torres, and Xavier Giro-i Nieto.
\newblock Recurrent neural networks for semantic instance segmentation.
\newblock \emph{arXiv preprint arXiv:1712.00617}, 2017.

\bibitem[SHI et~al.(2015)SHI, Chen, Wang, Yeung, Wong, and WOO]{NIPS2015_5955}
Xingjian SHI, Zhourong Chen, Hao Wang, Dit-Yan Yeung, Wai-kin Wong, and
  Wang-chun WOO.
\newblock Convolutional lstm network: A machine learning approach for
  precipitation nowcasting.
\newblock In C.~Cortes, N.~D. Lawrence, D.~D. Lee, M.~Sugiyama, and R.~Garnett,
  editors, \emph{Advances in Neural Information Processing Systems 28}, pages
  802--810. Curran Associates, Inc., 2015.
\newblock URL
  \url{http://papers.nips.cc/paper/5955-convolutional-lstm-network-a-machine-learning-approach-for-precipitation-nowcasting.pdf}.

\bibitem[Wang et~al.(2018)Wang, Ihler, Kording, and Yarkony]{Wang_2018_ECCV}
Shaofei Wang, Alexander Ihler, Konrad Kording, and Julian Yarkony.
\newblock Accelerating dynamic programs via nested benders decomposition with
  application to multi-person pose estimation.
\newblock In \emph{The European Conference on Computer Vision (ECCV)},
  September 2018.

\bibitem[Xiao et~al.(2018)Xiao, Wu, and Wei]{xiao2018simple}
Bin Xiao, Haiping Wu, and Yichen Wei.
\newblock Simple baselines for human pose estimation and tracking.
\newblock In \emph{European Conference on Computer Vision (ECCV)}, 2018.

\end{thebibliography}
\end{document}